\documentclass[10pt,twocolumn,letterpaper]{article}

\usepackage{iccv}
\usepackage{times}
\usepackage{epsfig}
\usepackage{graphicx}
\usepackage{amsmath}
\usepackage{amssymb}
\usepackage{multirow}
\usepackage{color}
\usepackage{caption}
\usepackage{booktabs}
\usepackage{subcaption}
\usepackage{cite}
\usepackage{enumitem}
\usepackage{mmstyle}
\usepackage{algorithm}
\usepackage{algorithmic}

\usepackage[pagebackref=true,breaklinks=true,letterpaper=true,colorlinks,bookmarks=false]{hyperref}

\iccvfinalcopy 

\def\iccvPaperID{1076} 

\ificcvfinal\fi
\begin{document}

\title{Towards Diverse and Natural Image Descriptions via a Conditional GAN}

\author{Bo Dai$^1$~~~~Sanja Fidler$^2$$^3$~~~~Raquel Urtasun$^2$$^3$$^4$~~~~Dahua Lin$^1$\\
$^1$Department of Information Engineering, The Chinese University of Hong Kong\\
$^2$University of Toronto
~~$^3$Vector Institute
~~$^4$Uber Advanced Technologies Group\\
{\footnotesize\texttt{db014@ie.cuhk.edu.hk}~~~~\texttt{fidler@cs.toronto.edu}~~~~\texttt{urtasun@cs.toronto.edu}~~~~\texttt{dhlin@ie.cuhk.edu.hk}}\\
}

\maketitle

\begin{abstract}
Despite the substantial progress in recent years, the 
image captioning techniques are still far from being perfect. 
Sentences produced by existing methods, \eg~those based on RNNs,
are often overly rigid and lacking in variability.
This issue is related to a learning principle widely
used in practice, that is, to maximize the likelihood of training samples.
This principle encourages high resemblance to the ``ground-truth'' captions,
while suppressing other reasonable descriptions.
Conventional evaluation metrics, \eg~BLEU and METEOR, also
favor such restrictive methods.
In this paper, we explore an alternative approach, with the aim to
improve the naturalness and diversity --
two essential properties of human expression.
Specifically, we propose a new framework based on
Conditional Generative Adversarial Networks (CGAN), which jointly
learns a generator to produce descriptions conditioned
on images and an evaluator to assess how well a description
fits the visual content.
It is noteworthy that training a sequence generator is
nontrivial. We overcome the difficulty by Policy Gradient,
a strategy stemming from Reinforcement Learning, which allows
the generator to receive early feedback along the way.
We tested our method on two large datasets, where
it performed competitively against real people in our user study
and outperformed other methods on various tasks.
\end{abstract}

\vspace{-5mm}

\section{Introduction}
\label{sec:intro}

\begin{figure}
	\centering
	\includegraphics[width=0.5\textwidth]{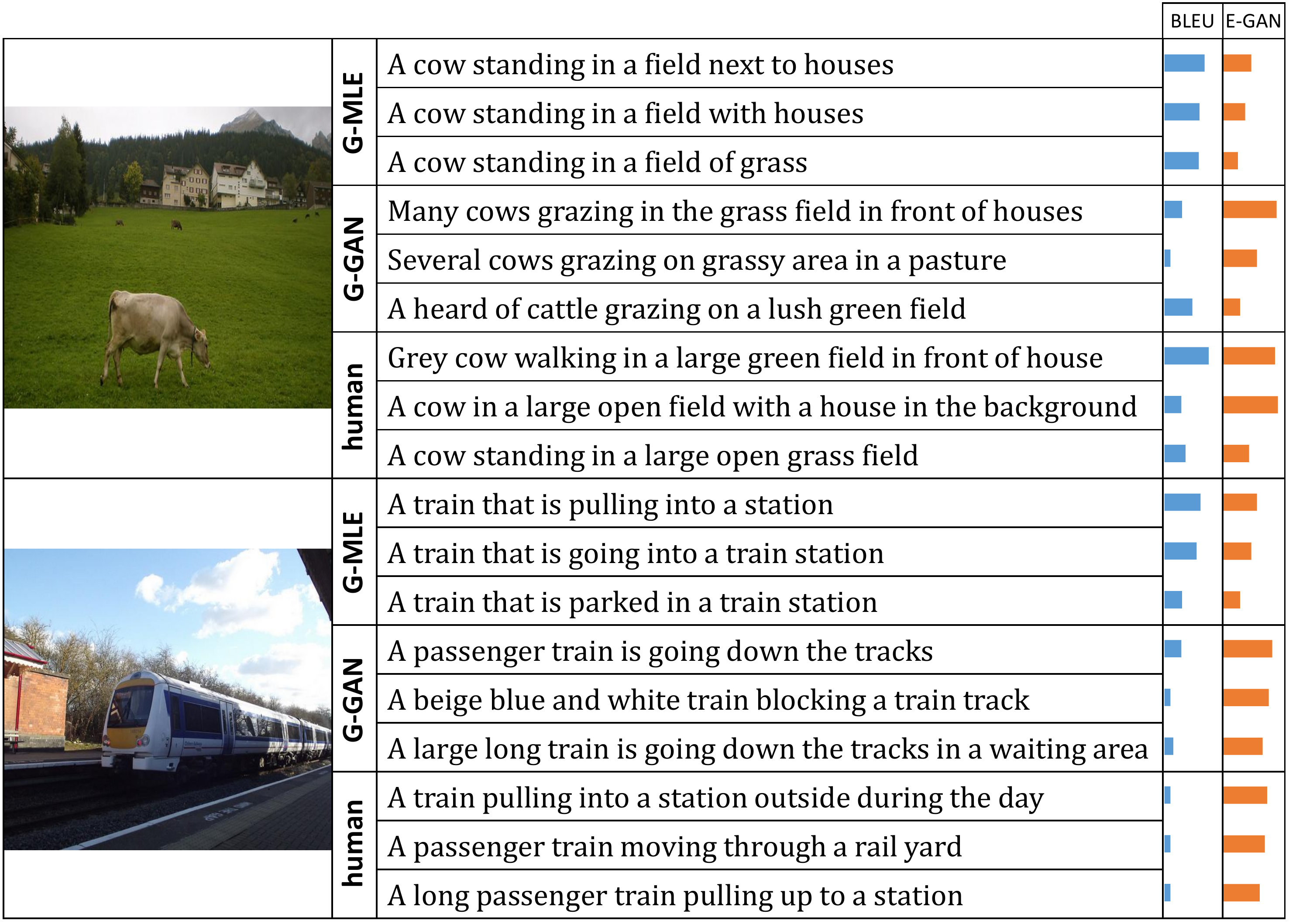}\\[-1mm]
	\caption{\small
	This figure shows two images with descriptions generated by humans,
	an LSTM net trained with our GAN-based framework (\emph{G-GAN}),
	and an LSTM net trained with MLE (\emph{G-MLE}).
	The last two columns compare the metric values of BLEU-3 and \emph{E-GAN},
	the evaluator trained using our method.
	As we can see, the sentences generated by \emph{G-GAN} are more natural
	and demonstrate higher variability, as compared to those by \emph{G-MLE}.
	Also, the \emph{E-GAN} metrics are more consistent with human's evaluations,
	while BLEU only favors those that significantly overlap
	with the training samples in detailed wording.}
	\label{fig:teaser_g}
	\vspace{-3.0mm}
\end{figure}


Generating descriptions of images has been an important task in
computer vision. Compared to other forms of semantic summary, \eg~object
tagging, linguistic descriptions are often richer, more comprehensive,
and a more natural way to convey image content. Along with the recent surge of deep learning technologies,
there has been remarkable progress in image captioning
over the past few years~\cite{vinyals2015show,xu2015show,you2016image,yang2016review,krause2016paragraphs}. Latest studies on this topic often
adopt a combination of an LSTM or its variant and a CNN. The former is
to produce the word sequences while the latter is
to capture the visual features of the images.


The advance in image captioning has been marked as a prominent
success of AI\footnote{ARTIFICIAL INTELLIGENCE AND LIFE IN 2030, \url{https://ai100.stanford.edu/2016-report}}.
It has been reported~\cite{vinyals2015show,xu2015show} that with certain
metrics, like BLEU \cite{papineni2002bleu}~or CIDEr \cite{vedantam2015cider}, state-of-the-art techniques
have already surpassed human's performance.
A natural question to ask is then: {\em has the problem of generating image
descriptions been solved?}
%
Let us take a step back, and look at a sample of the current results.
Figure~\ref{fig:teaser_g} shows
two vivid scenes together with three sentences produced by
the Encoder-and-Decoder model~\cite{vinyals2015show} (marked as ``G-MLE''), a state-of-the-art caption generator.
Though faithfully describing the content of the images,
these sentences feel rigid, dry, and lacking in vitality.

This is not surprising. Our brief survey (see Section~\ref{sec:relwork})
shows that existing efforts primarily focus on \emph{fidelity},
while other essential qualities of human languages,
\eg~\emph{naturalness} and \emph{diversity}, have received less attention.
More specifically, mainstream captioning models, including those
based on LSTMs~\cite{hochreiter1997long}, are mostly trained with the
(conditional) maximum likelihood objective. This objective
encourages the use of the \emph{n-grams} that appeared in the
training samples. Consequently, the generated sentences will
bear high resemblance to training sentences in \emph{detailed wording},
with very limited variability in expression~\cite{devlin2015exploring}.
Moreover, conventional evaluation metrics,
such as
BLEU \cite{papineni2002bleu},
METEOR \cite{lavie2014meteor},
ROUGE \cite{lin2004rouge}, and
CIDEr \cite{vedantam2015cider},
tend to favor this \emph{``safe''} but restricted way.
Under these metrics, sentences that contain matched n-grams
would get substantially higher scores than those using
variant expressions~\cite{anderson2016spice}.
This issue is manifested by the fact that human descriptions
get considerably lower scores.


Motivated to move beyond these limitations, we explore an alternative
approach in this work. We wish to produce sentences that possess
three properties:
(1) \textbf{Fidelity}: the generated descriptions should reflect the visual
content faithfully. Note that we desire the fidelity in \emph{semantics}
instead of \emph{wording}.
(2) \textbf{Naturalness}: the sentences should \emph{feel} like what real
people would say when presented with the image. In other words, when these
sentences are shown to a real person, she/he would ideally not be able to tell
that they are machine-generated.
(3) \textbf{Diversity}: the generator should be able to produce notably
different expressions given an image -- just like human beings, different
people would describe an image in different ways.


Towards this goal, we develop a new framework on top of the
Conditional GAN~\cite{mirza2014conditional}.
GAN has been successfully used in image generation.
As reported in previous works~\cite{reed2016generative,isola2016image}, they can produce \emph{natural}
images nearly indistinguishable from real photos,
freely or constrained by conditions.
This work studies a different task for the GAN method, namely,
generating \emph{natural} descriptions conditioned on a given image.
To our best knowledge, this is the first time the GAN method is
used for image description.


Applying GANs to text generation is nontrivial.
It comes with two significant challenges due to the special
nature of linguistic representation.
First, in contrast to image generation, where the transformation
from the input random vector to the produced image is a deterministic continuous
mapping, the process of generating a linguistic description is a
\emph{sequential sampling} procedure, which samples a \emph{discrete}
token at each step. Such operations are \emph{non-differentiable},
making it difficult to apply back-propagation directly.
We tackle this issue via \emph{Policy Gradient},
a classical method originating from reinforcement learning~\cite{sutton1999policy}.
The basic idea is to consider the production of each word
as an \emph{action}, for which the reward comes from
the evaluator. By approximating the stochastic policy
with a parametric function approximator, we allow gradients
to be back-propagated.

Second, in the conventional GAN setting, the generator
would receive feedback from the evaluator when an entire sample
is produced. For sequence generation, this would lead to several
difficulties in training, including \emph{vanishing gradients} and
\emph{error propagation}.
To mitigate such difficulties, we devise a mechanism that allows the
generator to get early feedback. Particularly, when a description
is \emph{partly} generated, our framework would calculate an approximated
\emph{expected future reward} through Monte Carlo rollouts~\cite{yu2016seqgan}.
Empirically, we found that this significantly improves the efficiency
and stability of the training process.

Overall, our contributions can be briefly summarized as follows:
(1) We explore an alternative approach to generate image descriptions,
which, unlike most of the previous work, encourages not only \emph{fidelity}
but also \emph{naturalness} and \emph{diversity}.
(2) From a technical standpoint, our approach relies on the conditional GAN
method to learn the generator, instead of using MLE, a paradigm
widely adopted in state-of-the-art methods.
(3) Our framework not only results in a generator that can produce
natural and diverse expressions, but also yields a description evaluator
at the same time, which, as we will show in our experiments, is
substantially more consistent with human evaluation.


\section{Related Work}
\label{sec:relwork}

\paragraph{Generation.}
Generating descriptions for images has been a long standing topic in computer vision.
Early studies mostly adopted \emph{detection-based} approaches.
Such methods first detect visual concepts (\eg~object categories, relationships, and attributes)
using CRFs~\cite{farhadi2010every, kulkarni2013babytalk, dai2017detecting},
SVMs~\cite{li2011composing}, or CNNs~\cite{fang2015captions, lisce2017},
then generate descriptions thereon using simple methods,
such as sentence templates~\cite{kulkarni2013babytalk,li2011composing},
or by retrieving relevant sentences from existing
data~\cite{farhadi2010every,fang2015captions,lebret2014simple,kuznetsova2012collective}.

In recent years,
the Encoder-and-Decoder paradigm proposed in~\cite{vinyals2015show}
became increasingly popular.
Many state-of-the-art frameworks~\cite{zhou2016image,you2016image,yang2016review,lu2016knowing,xu2015show,vinyals2015show} for this task adopt
the \emph{maximum likelihood} principle for learning.
Such a framework usually works as follows.
Given an image $I$, it first derives a feature representation $\vf(I)$,
and then generates the words $w_1, \ldots, w_T$ sequentially, following
a Markov process conditioned on $\vf(I)$.
The model parameters are learned via maximum likelihood estimation (MLE),
\ie~maximizing the conditional log-likelihood
of the training samples, as:
\begin{equation}
	\sum_{(I_i, S_i) \sim \cD}
	\sum_{t=0}^{T_i}
	\log p \left(w_i^{(t)} | \vf(I), w_i^{(t-1)}, \ldots, w_i^{(t-n)}\right)
\end{equation}
Here, $I_i$ and $S_i = (w_i^{(0)}, \ldots, w_i^{(T_i)})$ are the image and
the corresponding descriptive sentence of the $i$-th sample,
and $n$ is the order of the Markov chain
-- the distribution of the current word depends on $n$ preceding words.
Along with the popularity of deep neural networks, latest studies often
adopt neural networks for both image representation and language modeling.
For example, \cite{xu2015show} uses a CNN for deriving the visual features
$\vf(I)$, and an LSTM~\cite{hochreiter1997long}~net to express the sequential relations among words.
Despite the evolution of the modeling choices, the maximum likelihood principle
remains the predominant learning principle.

\begin{figure}
	\centering
	\includegraphics[height=0.22\textheight]{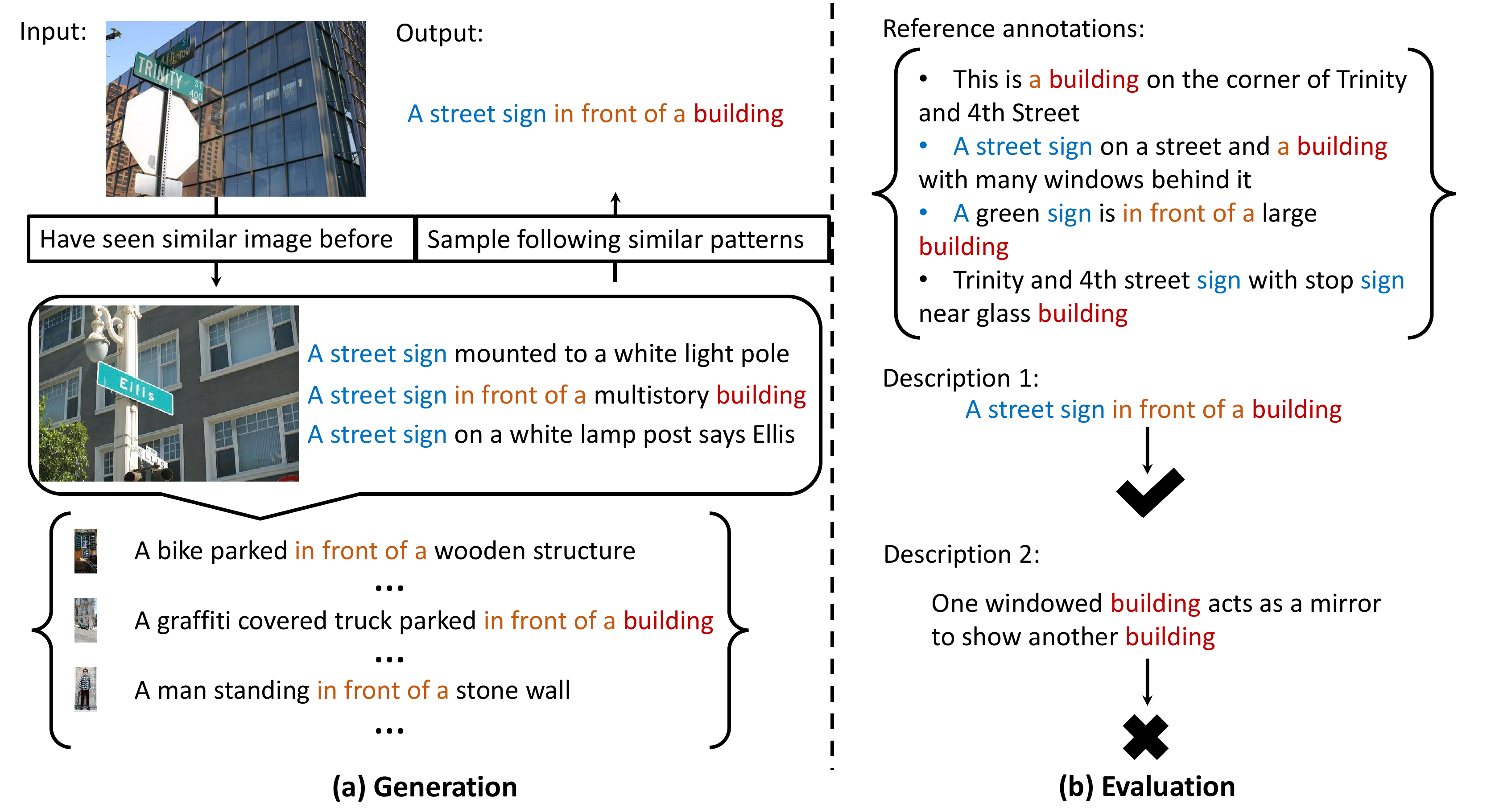}\\[-1.5mm]
	\caption{\small
		We illustrate the procedures of image description generation
		and evaluation for state-of-the-art approaches.
		While the generation procedure tends to follow observed patterns,
		the evaluation procedure also favors this point.
		Best viewed in color.}
	\label{fig:mle_limit}
\vspace{-1mm}
\end{figure}

\begin{figure}
	\centering
	\includegraphics[height=0.24\textheight]{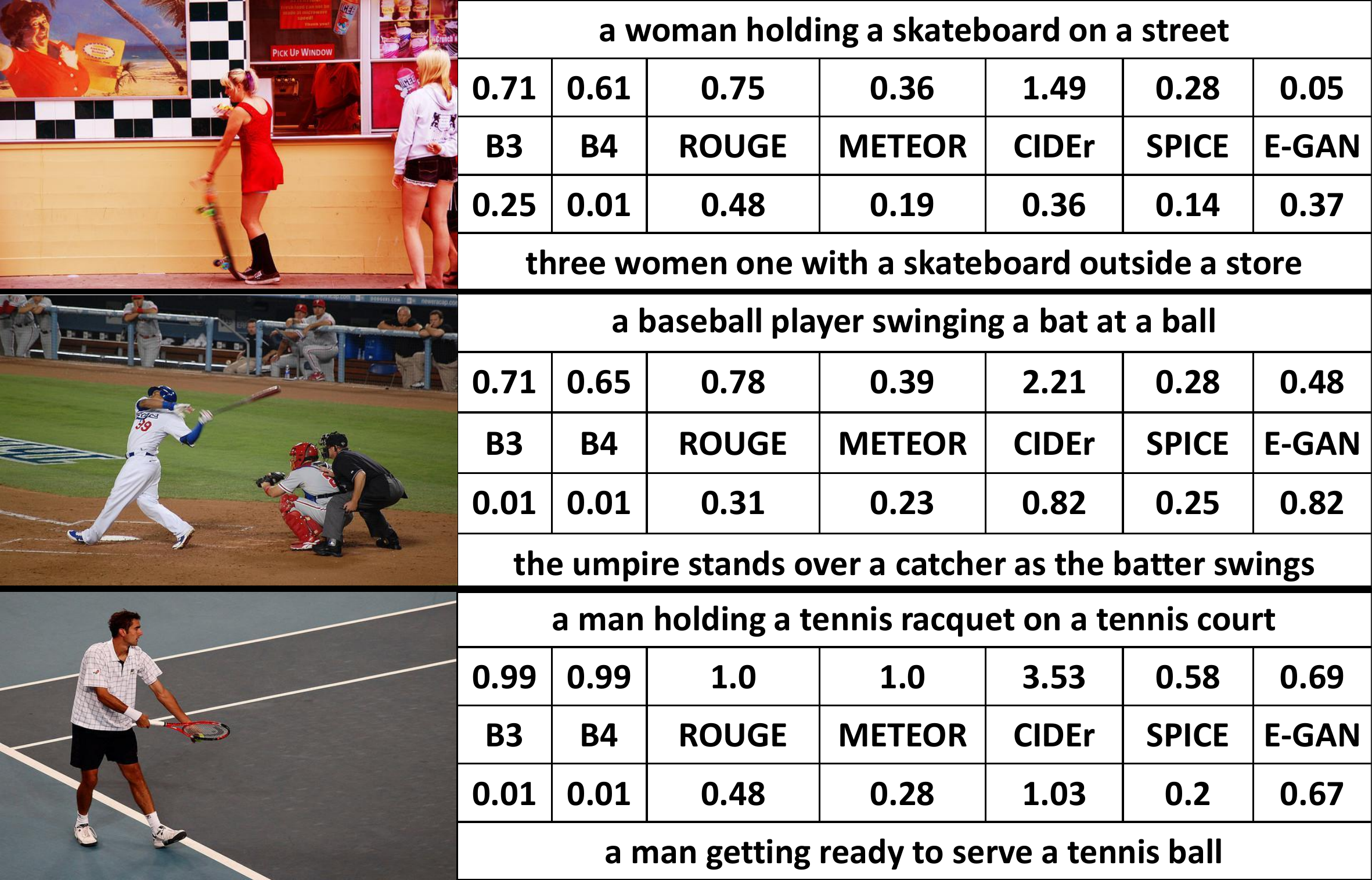}\\[-1.5mm]
	\caption{\small
	Examples of images with two semantically similar descriptions,
selected from ground-truth annotations.
While existing metrics assign higher scores to those with more matched n-grams,
	\emph{E-GAN} gives scores
	consistent with human evaluation.}
	\label{fig:oldmetrics}
	\vspace{-4mm}
\end{figure}

As illustrated in Figure~\ref{fig:mle_limit}, when similar images are presented,
the sentences generated by such a model often contain \emph{repeated} patterns \cite{devlin2015language}.
This is not surprising -- under the MLE principle, the joint probability of a sentence
is, to a large extent, determined by whether it contains the frequent n-grams from the training set. Therefore, the model trained in this way will tend
to produce such n-grams. In particular, when the generator yields a few of words
that match the prefix of a frequent n-gram, the remaining words of that
n-gram will likely be produced following the Markov chain.

\vspace{-12pt}
\paragraph{Evaluation.}
Along with the development of the generation methods,
various evaluation metrics have been proposed to assess
the quality of the generated sentences.
Classical metrics include BLEU~\cite{papineni2002bleu} and
ROUGE~\cite{lin2004rouge}, which respectively focuses on
the precision and recall of n-grams.
Beyond them, METEOR~\cite{lavie2014meteor} uses a combination of both the
precision and the recall of n-grams.
CIDEr\cite{vedantam2015cider} uses weighted statistics over n-grams.
As we can see, such metrics mostly rely on matching n-grams with
the \emph{``ground-truths''}.
As a result, sentences that contain frequent n-grams will get higher
scores as compared to those using variant expressions,
as shown in Figure~\ref{fig:oldmetrics}.
Recently, a new metric SPICE~\cite{anderson2016spice} was proposed.
Instead of matching between n-grams, it focuses on those linguistic
entities that reflect visual concepts (\eg~objects and relationships).
However, other qualities, \eg~the naturalness of the expressions,
are not considered in this metric.

\vspace{-12pt}
\paragraph{Our Alternative Way.}
Previous approaches, including both generation methods and evaluation metrics,
primarily focus on the \emph{resemblance} to the training samples.
While this is a \emph{safe} way to generate plausible descriptions,
it is \emph{limited}.
For example, when presented an image, different people would probably
give different descriptions that do not overlap much in the wording patterns.
This diversity in expression is an essential property of human languages,
which, however, is often overlooked in previous works (both generation
and evaluation).
In this work, we explore an alternative approach --
instead of emphasizing n-gram matching, we aim to improve the
\emph{naturalness} and \emph{diversity}, \ie~generating sentences that
feel like what real people would say, rather than focusing on word-by-word matching.
Specifically, our approach jointly trains a generator $G$ and an evaluator $E$
in an adversarial way, where $G$ is to produce natural descriptions,
while $E$ is to distinguish irrelevant or artificial descriptions
from natural ones.


From a technical standpoint, our approach is based on the conditional GAN approach.
GANs~\cite{goodfellow2014generative} and conditional GANs~\cite{mirza2014conditional} are
popular formulations for learning generators.
For computer vision, GAN was originally introduced to generate images~\cite{reed2016generative}.
In a recent work~\cite{yu2016seqgan}, a text generator based on the GAN method was proposed.
Note that this is an unconstrained generator that does not take into account any conditions.
Hence, it can not be directly used for generating descriptions for images
-- in this task, the relevance of the generated text to the given image is essential.
To our best knowledge, this is the first study that explores
the use of \emph{conditional} GAN in generating image descriptions.




\section{Framework}
\label{sec:frmwork}

\begin{figure*}[t]
\centering
\includegraphics[width=0.9\textwidth]{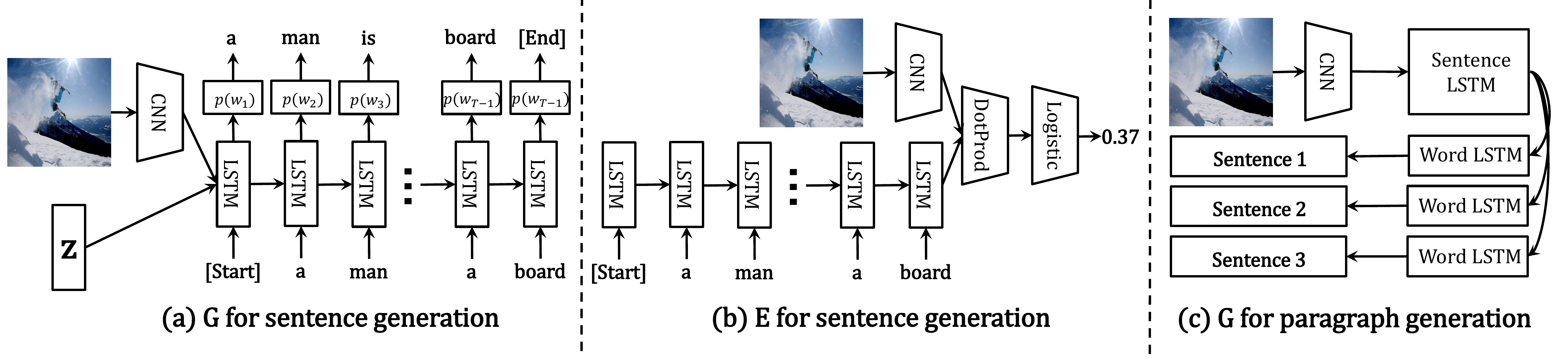}
\caption{\small
The structures of the generator $G$ for both single sentences and paragraphs,
and the evaluator $E$ for single sentences.}
\label{fig:structure}
\vspace{-3mm}
\end{figure*}

We propose a new framework for generating image descriptions based on
the conditional GAN~\cite{mirza2014conditional}~method, which consists
of a generator $G$, and an evaluator $E$.
Given an image $I$,
the former is for generating \emph{natural} and \emph{semantically relevant}
descriptions; while the latter is for evaluating how well a sentence or
paragraph describes $I$.
We start with generating single sentences as descriptions,
and then extend our framework to paragraph generation.

\subsection{Overall Formulation}

Our framework contains a \emph{generator} $G$ and a \emph{evaluator} $E$,
whose structures are respectively shown in Figure~\ref{fig:structure} (a) and (b).
It is worth noting that our framework is orthogonal to works
that focus on architectural designs of the $G$ and the $E$.
Their structures are not restricted to the ones introduced in this paper.
In our framework, given an image $I$, the generator $G$ takes two inputs:
an image feature $\vf(I)$ derived from a convolutional neural network
(CNN) and a random vector $\vz$.
In particular, we follow the setting in NeuralTalk2\footnote{\url{https://github.com/karpathy/neuraltalk2}},
adopting \emph{VGG16}~\cite{simonyan2014very} as the CNN architecture.
%
The random vector $\vz$ allows the generator to produce
different descriptions given an image. One can control the
\emph{diversity} by tuning the variance of $\vz$.
With both $\vf(I)$ and $\vz$ as the initial conditions,
the generator relies on an LSTM~\cite{hochreiter1997long}~net as a decoder,
which generates a sentence, word by word.
Particularly, the LSTM net assumes a sequence of latent states
$(s_0, s_1, \ldots)$. At each step $t$, a word $w_t$ is
drawn from the conditional distribution $p(w | s_t)$.

The evaluator $E$ is also a neural network, with an architecture
similar to $G$ but operating in a different way.
Given an image $I$ and a descriptive sentence $S = (w_0, w_1, \ldots)$,
it embeds them into vectors $\vf(I)$ and $\vh(S)$ of the same dimension,
respectively via a CNN and an LSTM net.
Then the \emph{quality} of the description, \ie~how well it describes $I$,
is measured by the dot product of the embedded vectors, as
\begin{equation}
    r_{\veta}(I, S) = \sigma \left(
        \langle \vf(I, \veta_I), \vh(S, \veta_S) \rangle
    \right).
\end{equation}
Here, $\veta = (\veta_I, \veta_S)$ denotes the evaluator parameters,
and $\sigma$ is a logistic function that turns the dot product
into a probability value in $[0, 1]$.
Note that while the CNN and the LSTM net in $E$ have the same structure
as those in $G$, their parameters are not tied with each other.

For this framework, the learning objective of $G$ is to
generate descriptions that are \emph{natural}, \ie~indistinguishable
from what humans would say when presented with the same image;
while the objective of $E$ is to distinguish between artifical
descriptions (\ie~those from $G$) and the real ones (\ie~those from
the training set).
This can be formalized into a minimax problem as follows:
\begin{equation}
    \min_{\vtheta} \max_{\veta}  \cL(G_{\vtheta}, E_{\veta}).
\end{equation}
Here, $G_{\vtheta}$ and $E_{\veta}$ are a generator with parameter $\vtheta$
and an evaluator with parameter $\veta$. The objective function $\cL$ is:
\begin{equation}\label{eq:minmax}
    \Ebb_{S \sim \cP_I} \left[
        \log r_{\veta}(I, S)
    \right]
    +
    \Ebb_{\vz \sim \cN_0} \left[
        \log(1 - r_{\veta}(I, G_{\vtheta}(I, \vz)))
    \right].
\end{equation}
Here, $\cP_I$ denotes the descriptive sentences for $I$
provided in the training set, $\cN_0$ denotes a standard
normal distribution, and $G_{\vtheta}(I, \vz)$ denotes
the sentence generated with $I$ and $\vz$.
The overall learning procedure alternates between the
updating of $G$ and $E$, until they reach an equilibrium.

This formulation reflects an essentially different philosophy
in \emph{how to train a description generator} as opposed to
those based on MLE. As mentioned, our approach aims at the
\emph{semantical relevance} and \emph{naturalness},
\ie~whether the generated descriptions feel like what human would say,
while the latter focuses more on word-by-word patterns.

\subsection{Training $G$: Policy Gradient \& Early Feedback}
\label{sec:g_train}

As mentioned, unlike in conventional GAN settings, the production
of sentences is a discrete sampling process, which is \emph{nondifferentiable}.
A question thus naturally arises - how can we \emph{back-propagate the
feedback} from $E$ under such a formulation?
We tackle this issue via \emph{Policy Gradient}~\cite{sutton1999policy}, a technique
originating from reinforcement learning.
The basic idea is to consider a sentence as a sequence of \emph{actions},
where each word $w_t$ is an action. The choices of such ``actions'' are
governed by a \emph{policy} $\vpi_{\vtheta}$.

With this interpretation, the generative procedure works as follows.
It begins with an empty sentence, denoted by $S_{1:0}$, as the initial state.
At each step $t$, the \emph{policy} $\vpi_{\vtheta}$ takes the conditions
$\vf(I)$, $\vz$, and the preceding words $S_{1:t-1}$ as inputs,
and yields a conditional distribution $\vpi_{\vtheta}(w_t | \vf(I), \vz, S_{1:t-1})$
over the extended vocabulary, namely all words plus an indicator of
sentence end, denoted by $e$. This computation is done by moving forward
along the LSTM net by one step.
From this conditional distribution, an action $w_t$ will be sampled.
If $w_t = e$, the sentence will be terminated, otherwise $w_t$ will
be appended to the end.
The \emph{reward} of this sequence of actions $S$ is $r_{\veta}(I, S)$,
the score given by the evaluator $E$.

Now, we have defined an action space, a policy, and a reward function,
and it seems that we are ready to apply the reinforcement learning method.
However, there is a serious issue here -- a sentence can only be evaluated
when it is \emph{completely} generated. In other words, we can only see the
reward at the end. We found empirically that this would lead to a number
of practical difficulties, \eg~gradients vanishing along a long chain
and overly slow convergence in training.

We address this issue through \emph{early feedback}. To be more specific,
we evaluate an \emph{expected future reward} as defined below
when the sentence is \emph{partially} generated:
\begin{equation}\label{eq:policy_value}
    V_{\vtheta, \veta}(I, \vz, S_{1:t}) =
    \Ebb_{S_{t+1:T} \sim G_{\vtheta}(I, \vz)}
    [r_{\veta}(I, S_{1:t}\oplus S_{t+1:T})].
\end{equation}
where $\oplus$ represents the concatenation operation.
Here, the expectation can be approximated using
Monte Carlo rollouts~\cite{yu2016seqgan}.
Particularly, when we have a part of the sentence $S_{1:t}$,
we can continue to sample the remaining words by simulating
the LSTM net until it sees an end indicator $e$. Conducting this
conditional simulation for $n$ times would result in $n$ sentences.
We can use the evaluation score averaged over these simulated sentences
to approximate the \emph{expected future reward}.
To learn the generator $G_{\vtheta}$, we use maximizing this expected reward
$V_{\vtheta,\veta}$ as the learning objective.
Following the argument in~\cite{sutton1999policy}, we can derive
the gradient of this objective \wrt~$\vtheta$ as:
\begin{equation} \label{eq:policy_grad}
    \tilde{\Ebb}\left[
        \sum_{t = 1}^{T_\text{max}}\sum_{w_t \in \cV}
        \nabla_{\vtheta} \pi_{\vtheta}(w_t|I, \vz, S_{1:t-1}) \cdot
        V_{\vtheta^\prime, \vpsi}(I, \vz, S_{1:t} \oplus w_t)
    \right].
\end{equation}
Here, $\cV$ is the vocabulary, $T_\text{max}$ is the max length of a description,
and $\tilde{\Ebb}$ is the mean over all simulated sentences
within a mini-batch.
$\vtheta'$ is a copy of the generator parameter $\vtheta$
at the begining of the update procedure of the generator.
During the procedure, the generator will be updated multiple times,
and each update will use the same set of parameters ($\vtheta'$) to compute Eq (\ref{eq:policy_value}).

Overall, using policy gradients, we make the generator trainable
with gradient descent.
Using expected future reward, we can provide early feedback
to the generator along the way, thus substantially improving
the effectiveness of the training process.
Note that policy gradients have also been used in image
description generation in \cite{rennie2016self,liu2016optimization}.
These works, however, adopt conventional metrics, \eg~BLEU and CIDEr
as rewards, instead of relying on GAN. 
Hence, their technical frameworks are fundamentally different.


\subsection{Training $E$: Naturalness \& Relevance}

The primary purpose of $E$ is to determine how well a description $S$
describes a given image $I$. A good description needs to satisfy
two criteria: \emph{natural} and \emph{semantically relevant}.
To enforce both criteria, inspired by \cite{reed2016generative} we extend Eq (\ref{eq:minmax}) to consider three types of descriptions
for each training image $I$:
(1) $\cS_I$: the set of descriptions for $I$ provided by human,
(2) $\cS_G$: those from the generator $G_{\vtheta}$, and
(3) $\cS_{\backslash I}$: the human descriptions for different images, 
which is uniformly sampled from all descriptions that are not associated with the given image $I$.
To increase the scores for the descriptions in $\cS_I$
while suppressing those in the others, we use a joint objective
formulated as:
\begin{equation}
    \max_{\veta} \quad \cL_E(\veta) = \frac{1}{N} \sum_{i=1}^N \cL_E(I_i;\veta).
\end{equation}
Here, $N$ is the number of training images. The term for each image $I_i$
is given by:
\begin{align}
    \cL_E (I;\veta)
    &= \Ebb_{S \in \cS_I} \log r_{\veta}(I, S) \notag \\
    &+ \alpha \cdot \Ebb_{S \in \cS_G} \log (1 - r_{\veta}(I, S)) \notag \\
    &+ \beta \cdot \Ebb_{S \in \cS_{\backslash I}} \log (1 - r_{\veta}(I, S)).
    \label{eq:e_loss}
\end{align}
The second term forces the evaluator to distinguish between the human descriptions
and the generated ones, which would in turn provide useful feedbacks to $G_{\vtheta}$,
pushing it to generate more \emph{natural} descriptions.
The third term, on the other hand, ensures the \emph{semantic relevance},
by explicitly suppressing mismatched descriptions.
The coefficients $\alpha$ and $\beta$ are to balance the contributions of these terms,
whose values are empirically determined on the validation set.

\subsection{Extensions for Generating Paragraphs}
\label{sec:paragen}
We also extend our framework to generate \emph{descriptive paragraphs}
by adopting a \emph{Hierarchical LSTM} design.
Specifically, our extended design is inspired by \cite{krause2016paragraphs}.
As shown in part (c) of Figure \ref{fig:structure}, it comprises two LSTM levels --
a \emph{sentence-level} LSTM net and a \emph{word-level} LSTM net.
Given the conditions $\vf(I)$ and $\vz$, to produce a paragraph,
it first generates a sequence of vectors based on $\vf(I)$, each encoding the topics of a sentence.
Then for each sentence, it generates the words conditioned on the
corresponding topic and the random vector $\vz$.

For evaluating a paragraph, the evaluator $E$ also adopts a hierarchical
design, but reversing the steps. Given an image $I$ and a paragraph $P$,
it first embeds each sentence into a vector via a word-level LSTM net, and
then embeds the entire paragraph by combining the sentence embeddings
via a sentence-level LSTM net.
Finally, it computes the score by taking the dot product between the paragraph
embedding $\vp$ and the image representation $\vf(I)$, and turning it into a
probability as $\sigma(\vp^T \vf(I))$, where
$\sigma$ is the logistic function.

After pretraining, we fix the sentence-level LSTM net of $G$ and only update the
word-level LSTM net of $G$ during the CGAN learning procedure.
This can effectively reduce the cost of Monte Carlo rollouts.
With a fixed sentence-level LSTM net, the policy gradients for each sentence
will be computed separately, following the steps in Sec~\ref{sec:g_train}.
Other parts of the training procedure remain the same.




\section{Experiment}
\label{sec:experiment}

\begin{table*}
\small
\centering
\begin{tabular}{|c||c|c|c|c|c|c|c||c|c|}
	\hline
	 			    & & BLEU-3 & BLEU-4 & METEOR & ROUGE\_L & CIDEr & SPICE & E-NGAN & E-GAN \\
	\hline
	\multirow{3}{*}{\rotatebox{90}{\small COCO}}
	& human                       & 0.290 & 0.192 & 0.240 & 0.465 & 0.849 & \textbf{0.211} & 0.527 & \textbf{0.626} \\
	& G-MLE                       & \textbf{0.393} & \textbf{0.299} & \textbf{0.248} & \textbf{0.527} & \textbf{1.020} & 0.199 & 0.464 & 0.427 \\
	& G-GAN 			    & 0.305 & 0.207 & 0.224 & 0.475 & 0.795 & 0.182 & \textbf{0.528} & 0.602 \\
	\hline
	\multirow{3}{*}{\rotatebox{90}{\small Flickr}}
	& human                       & 0.269 & 0.185 & 0.194 & 0.423 & 0.627 & 0.159 & 0.482 & \textbf{0.464} \\
	& G-MLE                       & \textbf{0.372} & \textbf{0.305} & \textbf{0.215} & \textbf{0.479} & \textbf{0.767} & \textbf{0.168} & 0.465 & 0.439 \\
	& G-GAN 			    & 0.153 & 0.088 & 0.132 & 0.330 & 0.202 & 0.087 & \textbf{0.582} & 0.456 \\
	\hline
\end{tabular}
\caption{This table lists the performances of different generators on MSCOCO and Flickr30k. 
On BLEU-\{3,4\}, METEOR, ROUGE\_L, CIDEr, and SPICE, \emph{G-MLE} is shown to be the best among all generators, 
surpassing human by a significant margin.
While \emph{E-NGAN} regard \emph{G-GAN} as the best generator, \emph{E-GAN} regard \emph{human} as the best one.}
\label{tab:sentgen}
\end{table*}

\begin{figure}
\centering
\includegraphics[height=0.15\textwidth]{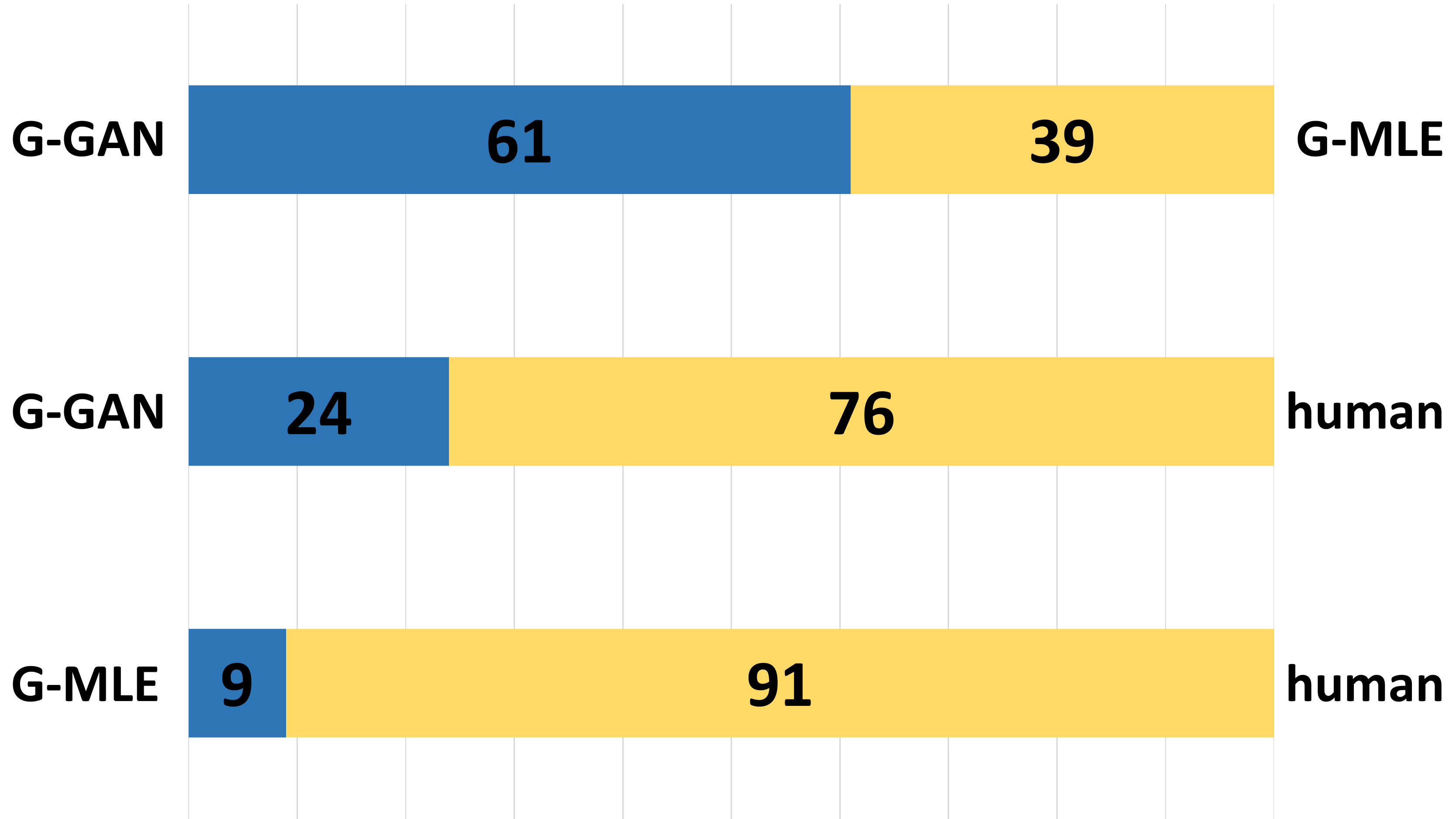}\\[-1.5mm]
\caption{\small
The figure shows the human comparison results between each pair of generators.
With names of the generators placed at each side of the comparison,
the blue and orange areas respectively indicate percentages of the generator in the left and right being the better one.}
\label{fig:userstudy}
\end{figure}

\begin{table}
\small
\centering
\begin{tabular}{|c|c|c|c|c|c|}
   \hline
	& & R@1 & R@3 & R@5 & R@10 \\
   \hline
   \multirow{2}{*}{S} 
   & G-MLE & 5.06 & 12.28 & 18.24 & 29.30 \\
   & G-GAN & \textbf{14.30} & \textbf{30.88} & \textbf{40.06} & \textbf{55.82} \\
   \hline
   \hline
   \multirow{2}{*}{P}
   & G-MLE & 9.88 & 20.12 & 27.30 & 39.94 \\
   & G-GAN & \textbf{12.04} & \textbf{23.88} & \textbf{30.70} & \textbf{41.78} \\
   \hline
\end{tabular}\\[-1.5mm]
\caption{\small
The recalls of image rankings for different generators.
Here recalls is the ratio of the original image being in the top-$k$ in the ranked lists.
The ranks are based on the similarities (S) between a image and a description,
estimated by \emph{E-GAN},
as well as the log-likelihoods (P), computed by different generators.}
\label{tab:imgrank}
\vspace{-3mm}
\end{table}

\begin{figure*}
\small
\centering
\includegraphics[width=0.9\textwidth]{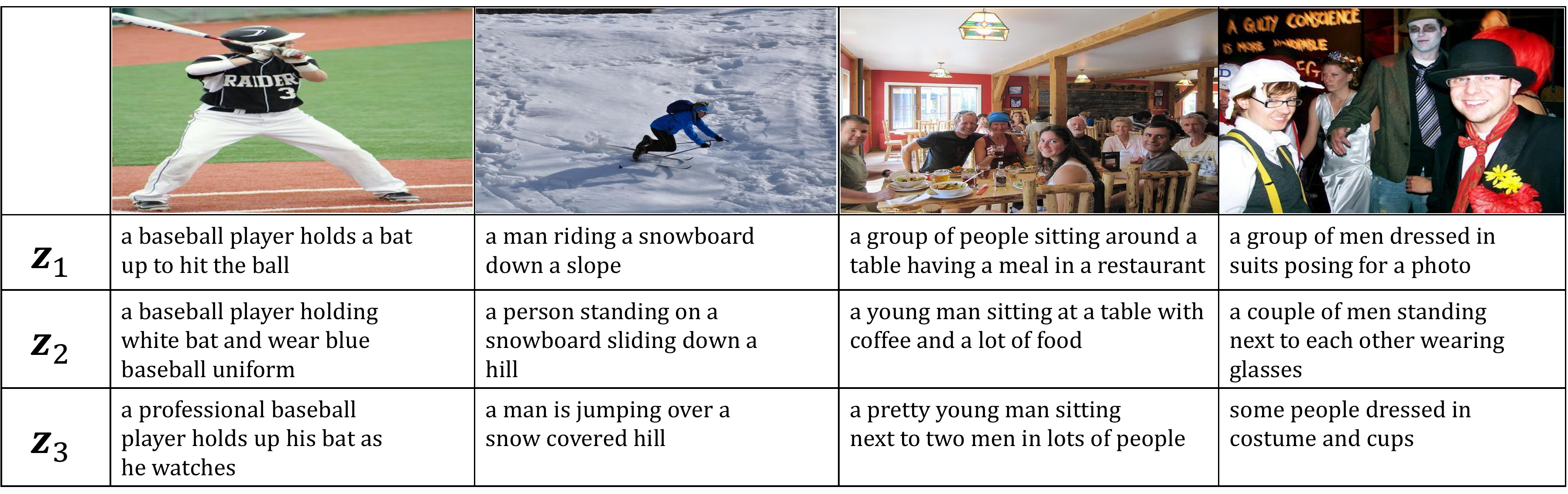}\\[-1.5mm]
\caption{This figure shows example images with descriptions generated by  \emph{G-GAN} with different $\vz$.}
\label{fig:noise}
\end{figure*}

\begin{figure*}
\small
\centering
\includegraphics[width=0.9\textwidth]{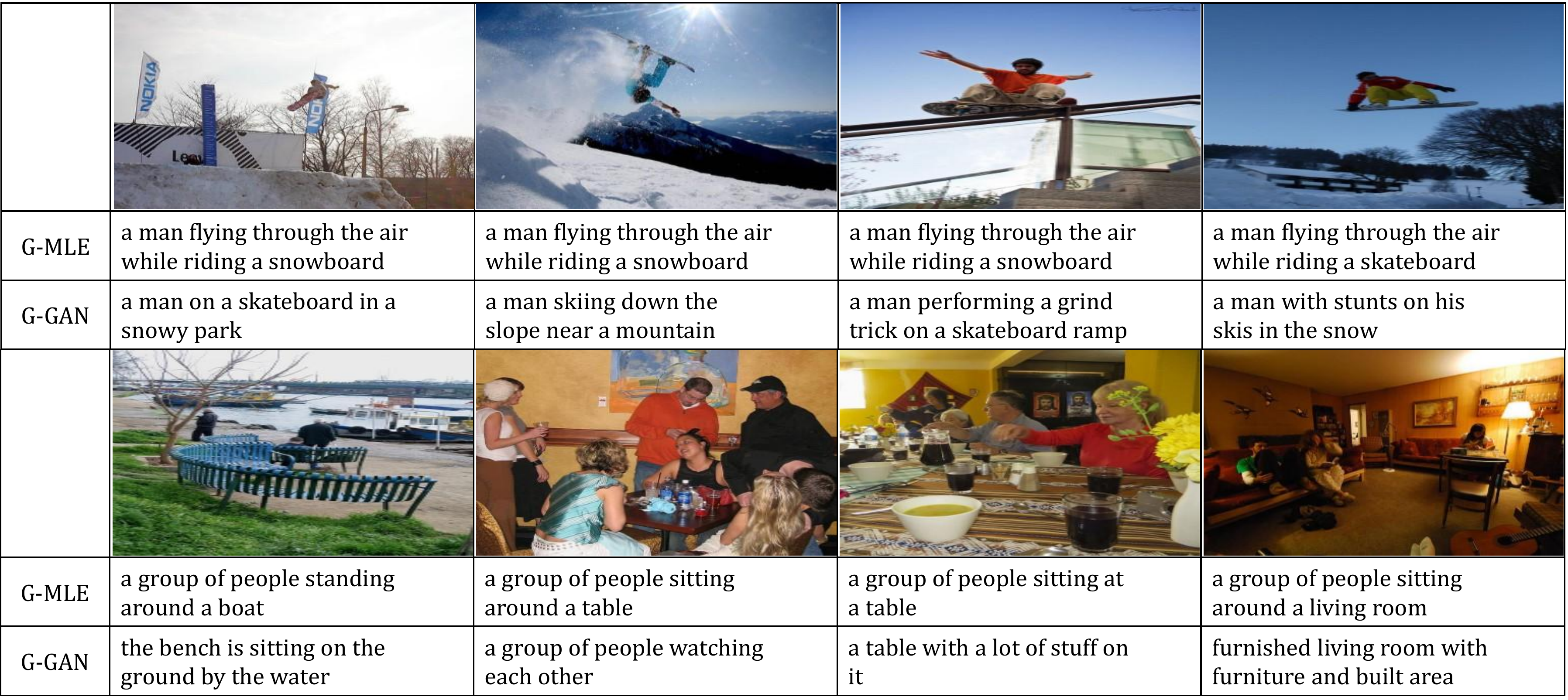}\\[-1.5mm]
\caption{This figure lists some images and corresponding descriptions generated by \emph{G-GAN} and \emph{G-MLE}. 
        \emph{G-MLE} tends to generate similar descriptions for similar images,
	while \emph{G-GAN} generates better distinguishable descriptions for them.}
\label{fig:compare}
\end{figure*}

\begin{figure*}
	\centering
	\includegraphics[width=0.9\textwidth]{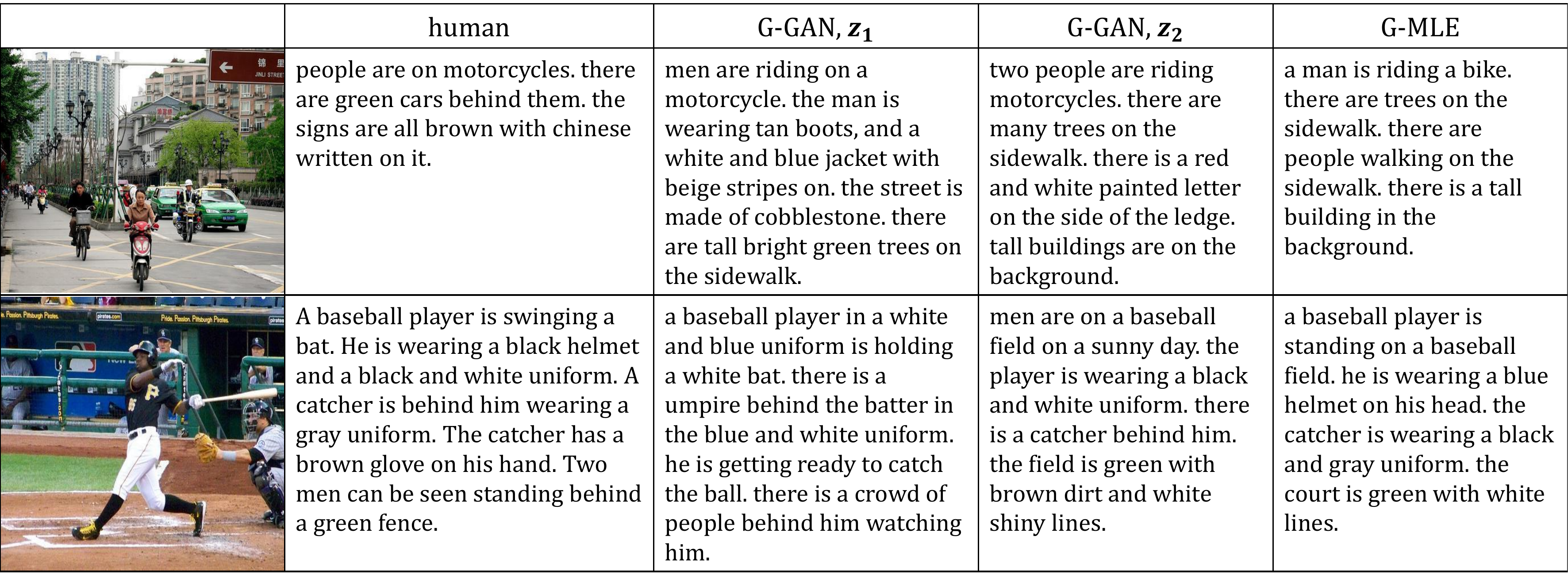}\\[-1.5mm]
	\caption{\small
	Examples of images with different descriptive paragraphs generated by a human,
	\emph{G-GAN} with different $\vz$, and \emph{G-MLE}.}
	\label{fig:paragen}
	\vspace{-5pt}
\end{figure*}

\paragraph{Datasets}
We conducted experiments to test the proposed framework
on two datasets:
(1) \emph{MSCOCO} \cite{lin2014microsoft},
which contains $82,081$ training images and $40,137$ validation images.
(2) \emph{Flickr30k} \cite{young2014image}, which contains $31,783$ images in total.
We followed the split in \cite{karpathy2015deep}, which has $1,000$ images for validation,
$1,000$ for testing, and the rest for training.
In both datasets, each image has at least $5$ ground truth sentences.
Note that our experiments involve comparison between human descriptions and
model-generated ones.
As we have no access to the ground-truth annotations of the testing images
in MSCOCO, for this dataset, we use the training set for both training and
validation, and the validation set for testing the performance.

\vspace{-11pt}
\paragraph{Experimental settings}
To process the annotations in each dataset,
we follow \cite{karpathy2015deep}~to remove non-alphabet characters,
convert all remaining characters to lower-case,
and replace all the words that appeared less than $5$ times with a special word \emph{UNK}.
As a result, we get a vocabulary of size $9,567$ on MSCOCO,
and a vocabulary of size $7,000$ on Flickr30k.
All sentences are truncated to contain at most $16$ words during training.
We respectively pretrain $G$ using standard MLE~\cite{vinyals2015show}, for $20$ epoches,
and $E$ with supervised training based on Eq~\eqref{eq:e_loss}, for $5$ epoches.
Subsequently, $G$ and $E$ are jointly trained, where
each iteration consists of one step of G-update followed by one step of E-update.
We set the mini-batch size to $64$, the learning rate to $0.0001$,
and $n=16$ in Monte Carlo rollouts.
When testing, we use beam search based on the expected rewards from E-GAN, 
instead of the log-likelihoods,
which we found empirically leads to better results. 

\vspace{-11pt}
\paragraph{Models}
We compare three methods for sentence generation:
(1)\textbf{Human}: a sentence randomly sampled from ground-truth annotations of each image
is used as the output of this method.
Other human-provided sentences will be used as the references for
metric evaluation.
This baseline is tested for the purpose of comparing human-provided
and model-generated descriptions.
(2)\textbf{G-MLE}: a generator trained based on MLE~\cite{vinyals2015show} is used
to produce the descriptions.
This baseline represents the state-of-the-art of mainstream methods.
(3)\textbf{G-GAN}: the same generator trained by our framework proposed in this paper,
which is based on the conditional GAN formulations.

For both \emph{G-MLE} and \emph{G-GAN},
\emph{VGG16}~\cite{simonyan2014very} is used as the image encoders.
Activations at the \texttt{fc7} layer, which are of dimension $4096$,
are used as the image features and fed to the description generators.
Note that \emph{G-GAN} also takes a random vector $\vz$ as input.
Here, $\vz$ is a $1024$-dimensional vector, whose entries are
sampled from a standard normal distribution.

\vspace{-11pt}
\paragraph{Evaluation metrics}
We consider multiple evaluation metrics, including six conventional metrics
BLEU-3 and BLEU-4\cite{papineni2002bleu},
METEOR\cite{lavie2014meteor},
ROUGE\_L\cite{lin2004rouge},
CIDEr\cite{vedantam2015cider},
SPICE\cite{anderson2016spice},
and two additional metrics relevant to our formulation:
\emph{E-NGAN} and \emph{E-GAN}.
Particularly,
\emph{E-GAN} refers to the evaluator trained using our framework,
\emph{E-NGAN} refers to the evaluator trained according to Eq~\eqref{eq:e_loss}
without updating the generator alternatively.
In other words, it is trained to distinguish between human-provided sentences
and those generated by an MLE-based model.

Table \ref{tab:sentgen} lists the performances of different generators under these metrics.
On both datasets, the sentences produced by \emph{G-MLE} receive considerably higher
scores than those provided by human, on nearly all conventional metrics.
This is not surprising. As discussed earlier, such metrics primarily focus on
n-gram matching \wrt~the references, while ignoring other important properties,
\eg~naturalness and diversity.
These results also clearly suggest that these metrics may not be particularly suited
when evaluating the overall quality of the generated sentences.
On the contrary, \emph{E-GAN} regards \emph{Human} as the best generator,
while \emph{E-NGAN} regards \emph{G-GAN} as the best one. These two metrics obviously take into account
more than just n-gram matching.

\vspace{-11pt}
\paragraph{User study \& qualitative comparison}
To fairly evaluate the quality of the generated sentences as well as
how \emph{consistent} the metrics are with human's perspective,
we conducted a user study.
Specifically, we invited $30$ human evaluators to compare the outputs
of different generators. Each time, a human evaluator would be
presented an image with two sentences from different methods
and asked to choose the better one. Totally, we collected
about $3,000$ responses.

The comparative results are shown in Figure \ref{fig:userstudy}:
From human's views,
\emph{G-GAN} is better than \emph{G-MLE} in $61\%$ of all cases.
In the comparison between human and models,
\emph{G-MLE} only won in $9\%$ of the cases, while
\emph{G-GAN} won in over $24\%$.
These results clearly suggest that the sentences produced
by \emph{G-GAN} are of considerably higher quality, \ie~being
more natural and semantically relevant.
The examples in Figure~\ref{fig:compare} also confirm
this assessment.
Particularly, we can see
when \emph{G-MLE} is presented with similar images,
it tends to generate descriptions that are almost the same.
On the contrary, \emph{G-GAN} describes them with more distinctive and diverse ones.
We also varied $\vz$ to study the capability of \emph{G-GAN}
in giving \emph{diverse} descriptions while maintaining the semantical relatedness.
The qualitative results are listed in Figure~\ref{fig:noise}.

For the evaluation metrics, the assessments provided by \emph{E-GAN}
are the most consistent with human's evaluation, 
where the Kendall's rank correlation coefficient between \emph{E-GAN} and \emph{HE} is 0.14,
while that for CIDEr and SPICE are -0.30 and -0.25.
%
Also note that \emph{E-GAN} yields a larger numerical gap
between scores of human and those of other generators as compared to \emph{E-NGAN},
which suggests that
adversarial training can improve the discriminative power of the evaluator.

\vspace{-15pt}
\paragraph{Evaluation by retrieval}

To compare the \emph{semantic relevance}, we conducted an experiment
using generated descriptions for retrieval.
Specifically, we randomly select $5,000$ images from the MSCOCO validation set;
and for each image, we use the generated description as a query,
ranking all $5,000$ images according to the similarities between the images and the descriptions,
computed by \emph{E-GAN},
as well as the log-likelihoods.
Finally, we compute the recall of the original image that appeared in the top-$k$ ranks.
The results for $k=1, 3, 5, 10$ are listed in Table \ref{tab:imgrank},
where \emph{G-GAN} is shown to provide more discriminative descriptions,
outperforming \emph{G-MLE} by a large margin across all cases.


\vspace{-15pt}
\paragraph{Failure Analysis}
We analyzed failure cases and found that a major kind of errors is 
the inclusion of incorrect details. \eg colors (red/yellow hat),
and counts (three/four people). 
A possible cause is that there are only a few samples for each particular detail,
and they are not enough to make the generator capture these details reliably.
Also, the focus on diversity and overall quality may also encourage the generator to include more details, 
with the risk of some details being incorrect.

\vspace{-11pt}
\paragraph{Paragraph Generation}
We also tested our framework on paragraph generation (See Sec~\ref{sec:paragen}).
We use the dataset provided by \cite{krause2016paragraphs},
which contains $14,575$ training images, $2,487$ validation images, and $2,489$ testing images.
Example results are shown in Figure \ref{fig:paragen}.
Again, we found that G-GAN can produce diverse and more natural descriptions as compared
to G-MLE, which tends to follow similar patterns across sentences.

\section{Conclusion}
\label{sec:concls}

This paper presented an alternative approach to generating image descriptions.
Compared to existing methods, which are mostly focused on the match of detailed wording,
our approach, instead, aims to improve the overall quality,
which involves \emph{semantic relevance}, \emph{naturalness}, and \emph{diversity}.
Some of these properties are often overlooked in previous efforts.
We proposed a formulation based on conditional GAN that jointly
trains a generator $G$ and an evaluator $E$, and applied Policy Gradient and
early feedbacks to tackle the technical challenges in end-to-end training.
On both MSCOCO and Flickr30k, the proposed method produced descriptions
that are more natural, diverse, and semantically relevant as compared
to a state-of-the-art MLE-based model. This is clearly
demonstrated in our user studies, qualitative examples, and retrieval applications.
Our framework also provides an evaluator that is more consistent with human's evaluation.

\vspace{-13pt}
\paragraph{Acknowledgment}
This work is partially supported by
the Big Data Collaboration Research grant from SenseTime Group (CUHK Agreement No.TS1610626), 
the General Research Fund (GRF) of Hong Kong (No.14236516)
and the Early Career Scheme (ECS) of Hong Kong (No.24204215).

{\small
\bibliographystyle{ieee}
\bibliography{gantext}
}

\end{document}